\documentclass[10pt,twocolumn,letterpaper]{article}

\usepackage[final]{cvpr}      
\usepackage{algorithm}
\usepackage{algorithmic}
\usepackage{amsmath,amssymb,amsfonts}
\usepackage{bm}
\usepackage{booktabs}
\usepackage{caption}
\usepackage{colortbl}
\usepackage{graphicx}
\usepackage{multirow, multicol} 
\usepackage[table]{xcolor} 
\usepackage[switch]{lineno}

\usepackage[accsupp]{axessibility} 

\definecolor{cvprblue}{rgb}{0.21,0.49,0.74}
\usepackage[pagebackref,breaklinks,colorlinks,allcolors=cvprblue]{hyperref}

\def\paperID{2577} 
\def\confName{CVPR}
\def\confYear{2026}

\title{Reinforcement-Guided Synthetic Data Generation for Privacy-Sensitive \\Identity Recognition}

\author{Xuemei Jia$^{1,2}$~~~ Jiawei Du$^{3}$~~~ Hui Wei$^{4}$~~~ Jun Chen$^{1,2\dag}$~~~ Joey Tianyi Zhou$^{3}$~~~ Zheng Wang$^{1,2\dag}$ \\
\small$^{1}$National Engineering Research Center for Multimedia Software, School of Computer Science, Wuhan University, Wuhan, China~~~\\
\small$^{2}$Hubei Key Laboratory of Multimedia and Network Communication Engineering, Wuhan, China\\
\small$^{3}$Centre for Frontier AI Research (CFAR) \& Institute of High Performance Computing (IHPC), A*STAR, Singapore\\
\small$^{4}$Center for Machine Vision and Signal Analysis (CMVS), University of Oulu, Finland\\
{\tt\small \{jiaxuemeil,chenj,wangzwhu\}@whu.edu.cn, \{dujw,Joey\_Zhou\}@cfar.a-star.edu.sg,}\\
{\tt\small huiwei.truth@gmail.com}
}

\begin{document}
\maketitle

{
\renewcommand\thefootnote{}
\footnotetext{$^\dag$Corresponding author}
}

\begin{abstract}
High-fidelity generative models are increasingly needed in privacy-sensitive scenarios, where access to data is severely restricted due to regulatory and copyright constraints. This scarcity hampers model development—ironically, in settings where generative models are most needed to compensate for the lack of data. This creates a self-reinforcing challenge: limited data leads to poor generative models, which in turn fail to mitigate data scarcity. 
To break this cycle, we propose a reinforcement-guided synthetic data generation framework that adapts general-domain generative priors to privacy-sensitive identity recognition tasks.
We first perform a cold-start adaptation to align a pretrained generator with the target domain, establishing semantic relevance and initial fidelity.
Building on this foundation, we introduce a multi-objective reward that jointly optimizes semantic consistency, coverage diversity, and expression richness, guiding the generator to produce both realistic and task-effective samples.
During downstream training, a dynamic sample selection mechanism further prioritizes high-utility synthetic samples, enabling adaptive data scaling and improved domain alignment.
Extensive experiments on benchmark datasets demonstrate that our framework significantly improves both generation fidelity and classification accuracy, while also exhibiting strong generalization to novel categories in small-data regimes.

\end{abstract}
    
\section{Introduction}
\label{sec:intro}

High-performing visual models hinge on abundant and well-annotated data. Yet privacy regulations, annotation costs, and copyright constraints make large-scale collection increasingly difficult, throttling generalization in practice~\cite{amershi2014power,voigt2017eu,du2023minimizing}. 
These limitations hinder the development of deep learning models and restrict their generalization ability. Synthetic data generation offers a promising alternative~\cite{nips/GoodfellowPMXWOCB14,nips/HoJA20,cvpr/RombachBLEO22,wang2025scone}, but its success still depends on the availability of high-quality real data. In data-scarce settings, the absence of sufficient supervision often leads to low-fidelity samples and poor task utility, creating a vicious cycle: insufficient real data produces weak synthetic data, which in turn fails to mitigate the original data scarcity.

\begin{figure}   
    \centering
    \includegraphics[width=0.85\columnwidth]{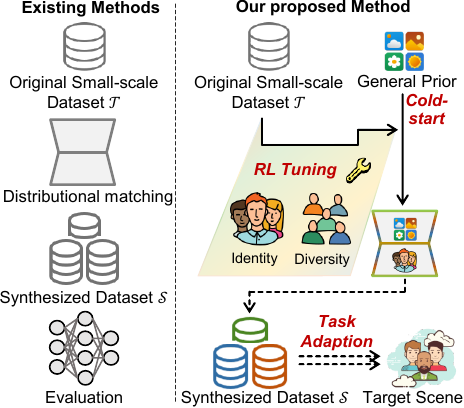}  
    \caption{Pipeline comparison. (a) Existing methods rely solely on specific data, resulting in limited diversity and low utility of synthesized images. (b) We adapt broad, general-domain priors to the target domain, improving both diversity and task utility.}  
    \vspace{-5mm}
    \label{motiv}
\end{figure}

Many efforts have sought to improve the utility of synthetic data. A straightforward direction is data augmentation~\cite{aaai/Zhong0KL020}. For example, \citet{karras2020training} employed adaptive augmentation to increase variation under limited data, while \cite{cvpr/Sun019,tcsvt/ZhangLZW24} explored manipulating virtual environments to enrich diversity.
Although these methods expand appearance variety, they struggle to close the domain gap between synthetic and real images.
With the emergence of GANs~\cite{nips/GoodfellowPMXWOCB14} and diffusion models~\cite{nips/HoJA20}, subsequent work attempted to enhance data diversity by factorizing visual attributes such as identity, pose, and background.
Approaches like \cite{zheng2019joint,kolf2023identity,miao2024training} decomposed and recombined visual components to synthesize new instances, yet the improvement in downstream performance remained limited.
Overall, existing methods primarily aim to match the distribution of real data, effectively mimicking the source distribution rather than surpassing it in task utility, as shown in Figure~\ref{motiv}.

Motivated by this observation, we advocate a paradigm shift: instead of relying solely on the scarce source distribution, we leverage rich general-domain priors as guidance for synthesis.
Pre-trained models encompassing large-scale vision backbones and generative architectures trained on diverse datasets encode structural, semantic, and contextual knowledge that complements the scarcity of domain-specific data. Incorporating such priors provides a broader representational basis and allows synthetic data to exhibit richer diversity and higher fidelity, even when direct supervision is limited.

The central challenge lies in effectively adapting general priors to target domains with limited data availability. In response, we formulate the synthesis process as a reinforcement learning problem, where the generative model acts as a policy that produces synthetic samples and receives rewards according to their contribution to downstream tasks. This reward-guided formulation enables task-aligned adaptation guided by performance feedback rather than direct supervision. Using proxy objectives, the generator learns to produce samples that are both visually plausible and functionally relevant to the target domain. Through this adaptive loop, the model progressively bridges the gap between general priors and domain-specific demands.

Building on this insight, our approach forms a continuous adaptive process rather than three isolated stages.
We begin with a cold-start adaptation that aligns a pre-trained generator with the target domain, establishing semantic consistency and basic fidelity as a foundation for further optimization.
On top of this initialization, we progressively refine the generator through a task-specific reinforcement signal.
A multi-component reward, capturing semantic alignment, coverage diversity, and expression richness, guides the model to generate data that are both visually realistic and task-effective.
This adaptation not only enhances synthesis quality but also yields an internal measure of sample utility.
During downstream task training, we extend this principle through a dynamic sample selection mechanism, which leverages the learned utility signals to emphasize task-relevant samples and mitigate distributional bias.

In a nutshell, our contribution can be summarized as: 
\begin{itemize}
    \item We identify and address the fundamental limitation of existing generative methods, where low-data supervision perpetuates a data scarcity loop. We propose a reinforcement-guided synthesis framework that leverages general-domain priors to mitigate this issue.
    \item We propose a task-specific reward that balances fidelity, diversity, and task relevance, guiding the generator toward utility-driven synthesis.
    \item We design a dynamic sample selection strategy that further improves generalization under domain shift, achieving consistent gains across privacy-sensitive benchmarks.
\end{itemize}
\section{Related Works}
\label{rw}

\subsection{Data Scarcity in Identity-related Tasks}
Collecting and annotating real-world datasets is costly and time-consuming. In privacy-sensitive domains such as identity recognition, these challenges are further intensified by legal and ethical restrictions that limit large-scale data acquisition and sharing~\cite{wacv/KansalWK24,jia2025balancing,hu2025contrastive}. To alleviate data scarcity, prior studies have explored both augmentation and synthesis strategies. 
Early methods relied on conventional augmentations, such as random resizing, cropping, flipping~\cite{tmm/LuoJGLLLG20}, erasing~\cite{aaai/Zhong0KL020}, and their combinations~\cite{karras2020training}, to enhance intra-dataset diversity. Beyond these, synthetic human images have been generated through controllable virtual environments (e.g., GTA V)~\cite{cvpr/Sun019,corr/abs-2211-11165,tcsvt/ZhangLZW24}.

The rise of generative models such as GANs~\cite{nips/GoodfellowPMXWOCB14} and diffusion models~\cite{nips/HoJA20,zhong2023refined} has greatly advanced realistic person image synthesis~\cite{lin2023privacy,tip/JiaZYLH22,corr/abs-2406-06045,du2024diversity}. These efforts generally fall into two paradigms: (i) disentangling and recombining factors across real samples~\cite{zheng2019joint,wu2023uncovering}, and (ii) conditional generation guided by attributes or control signals~\cite{mirza2014conditional,zhang2023adding}.
Despite these advances, their performance is still constrained by the limited scale, diversity, and representational richness of available data, an issue that becomes even more pronounced in privacy-constrained domains.

\subsection{Conditional Diffusion Models}
Diffusion models have emerged as a powerful class of generative models, achieving SOTA performance in high-fidelity image synthesis. Early methods, {e.g.} DDPM~\cite{nips/HoJA20}, formulate a generative process through a Markovian forward diffusion and a learned reverse denoising process, while subsequent variants~\cite{icml/NicholD21,song2020denoising}, improve sampling efficiency and quality.
To further reduce computational costs, Latent Diffusion Models (LDMs)~\cite{cvpr/RombachBLEO22} perform diffusion in a compact latent space, preserving semantic fidelity while enabling large-scale text-guided generation, as demonstrated by Imagen~\cite{nips/SahariaCSLWDGLA22} and Stable Diffusion~\cite{cvpr/RombachBLEO22}.

However, convolution-based U-Nets in LDMs are inherently limited in capturing long-range dependencies and global semantic coherence.
To address it, Diffusion Transformers~\cite{iccv/PeeblesX23} (DiT) replace U-Net with a Transformer-based backbone, enhancing global context modeling and long-range interaction.
This makes DiT particularly suitable for purely visual generation in the latent space without textual prompts, while preserving global semantic consistency.

\subsection{RL-based Diffusion Model Optimization}
Recent studies have reformulated the diffusion denoising process as a sequential decision-making problem under the reinforcement learning (RL) framework~\cite{wallace2024diffusion}, enabling the model to optimize sample quality through reward-driven policy updates rather than fixed likelihood objectives.
~\cite{fan2023dpok} introduces DPOK, an online RL framework with KL-regularized policy optimization for fine-tuning text-to-image diffusion models, achieving superior text–image alignment and image fidelity compared to reward-weighted supervised fine-tuning.
~\cite{han2024advancing} develops an RL framework with comparative feedback and adaptive condition embeddings for accurate and consistent report-conditioned chest X-ray generation.
~\cite{saremi2025rl4med} uses attribute recognition accuracy as a reward signal to guide policy optimization in medical image synthesis.
~\cite{miao2024training} proposes a diversity-oriented RL fine-tuning framework that enhances generative diversity but still depends on diverse, unbiased reference images.

In contrast, our work targets privacy-constrained domains where large-scale data collection is infeasible. We emphasize learning robust and generalizable feature representations from restricted distributions, while ensuring the generated data remain diverse and semantically meaningful.

\section{Preliminary}
\label{pre}
\subsection{Diffusion Transformer}
Diffusion Transformers (DiT)~\cite{iccv/PeeblesX23} excel at modeling global context and long-range spatial dependencies, enabling semantically coherent image generation without relying on textual prompts. It consists of two core components:
\textit{Autoencoder.} The encoder $E$ maps the input image $x$ to a latent representation $z = E(x)$, and the decoder $D$ reconstructs the image as $\hat{x} = D(z)$. This design enables the diffusion process to operate in a compressed yet semantically rich latent space.
\textit{Latent Diffusion Transformer.} The latent code is first patchified into a sequence of visual tokens which are then processed by a stack of DiT blocks. Each block applies self-attention and feed-forward layers, conditioned on the diffusion timestep and an optional class embedding, resulting in an intermediate output. Finally, the processed tokens are reassembled into a refined latent feature map.

\subsection{RL finetuned Diffusion}
The iterative denoising process of diffusion probabilistic models can be formulated as a multi-step Markov decision process, where each denoising step corresponds to a state transition governed by the model policy. Through this sequential refinement, the model progressively transforms a noisy input into a high-quality sample.

In this formulation, conditional diffusion can be optimized via reinforcement learning, where the objective is to maximize the expected reward of generated samples given specific conditions:
$J_\theta = \mathbb{E}_{p(c)} \left[ \mathbb{E}_{p_\theta(x|c)}\bigl[R(x, c)\bigr] \right]$. $p(c)$ is the distribution over conditioning labels, 
$p_{\theta}(x|c)$ denotes the sample distribution generated by a pretrained model under condition $c$, and $R(x,c)$ is a task-specific reward function that evaluates the fidelity, utility, or realism of the generated image $x$. The reward is typically calculated based on the terminal sample $x_0$ produced by the diffusion trajectory.

Following the DPOK framework~\cite{fan2023dpok}, the gradient of the objective with respect to the model parameters \(\theta\) can be derived using the policy gradient theorem: 
\begin{equation}
\nabla_\theta J_\theta = \mathbb{E}_{x_{1:T}}\left[ R(x, c) \sum_{t} \nabla_\theta \log p_\theta\bigl(x_{t-1} \mid x_t, c, t\bigr) \right],
\label{eq:policy_gradient}
\end{equation}
where $T$ is the total number of timesteps in the diffusion process, and $x_t$ denotes the intermediate state at timestep $t$. Each gradient term optimizes the denoising policy to produce samples that yield higher rewards.
By iteratively applying this policy gradient update, the diffusion model can be fine-tuned to enhance image fidelity, structural details, and semantic alignment with the conditioning signals, even when only limited supervision or feedback is available.

\section{Proposed Method}
We tackle the challenge of data scarcity in privacy-sensitive domains by adapting publicly available generative models pretrained on general-domain data to target tasks with limited supervision. Our framework consists of three sequential stages.
First, a cold-start adaptation stage aligns the pretrained generator with the target distribution, establishing a strong initialization for downstream optimization.
Second, a reinforcement learning-based fine-tuning stage leverages a task-specific reward function to guide the generator toward producing samples with higher fidelity, diversity, and task relevance.
Finally, a dynamic sample selection mechanism prioritizes task-relevant samples to enhance generalization under distribution shifts.
Collectively, these components constitute a flexible and effective pipeline for generating utility-preserving data in privacy-constrained settings.

\subsection{Cold-Start Initialization}

To enable effective adaptation of pre-trained diffusion models to new domains with limited data, we introduce a cold-start initialization protocol as the first step of our framework. Specifically, given a publicly available DiT~\cite{iccv/PeeblesX23} pre-trained on large-scale generic datasets such as ImageNet~\cite{cvpr/DengDSLL009}, we define the initialization step as 
$\theta_0=\text{Init}(\theta_\text{pre},X)$, where $\theta_\text{pre}$ denotes the pre-trained model parameters and $X$ represents the target-domain dataset. The function 
$\text{Init}(\cdot)$ performs lightweight fine-tuning to produce $\theta_0$, which serves as a stable and semantically aligned starting point for subsequent reward-guided refinement.

In practice, we replace the class embedding of the pretrained DiT with a task-specific head aligned to the target label space. Fine-tuning is performed using a standard denoising objective on limited target samples while keeping the backbone frozen. Only a few hyperparameters, such as the learning rate and iteration number, are adjusted to mitigate overfitting. This minimal adaptation preserves the generalization ability of the pre-trained model while introducing task-relevant inductive bias into the generation process.

This initialization step is particularly helpful under privacy-constrained conditions, where direct reward optimization can be unstable due to data scarcity. By providing a well-aligned initialization, the model achieves more reliable convergence in the next reinforcement learning stage.

\subsection{Reward-driven Optimization}
\label{sec:reward}
While the cold-start initialization provides a stable and semantically aligned starting point, it does not explicitly enforce the generation of identity-relevant or diverse samples. To further refine the model toward these objectives, we introduce a reinforcement learning-based optimization stage.

In this stage, a reward function is introduced to guide the conditional diffusion process toward identity-preserving and semantically meaningful image generation. The reward comprises three components: \textit{semantic consistency}, which enforces alignment between generated and reference representations; \textit{distributional coverage}, which encourages coverage of target-domain variability; and \textit{expressive diversity}, which promotes visually diverse yet coherent samples. These components jointly drive the diffusion model toward achieving both discriminative relevance and generative richness in privacy-constrained settings.

\subsubsection{Semantic Consistency}
To preserve identity information during generation, we measure semantic consistency in the feature space, ensuring that the generated representation remains close to the semantic center of its corresponding identity.
Let $\mathcal{B}_y=\{f_i\}_{i=1}^{N_y}$ denote the set of reference features stored in a memory bank for identity $y$.
The class prototype is computed as the mean-normalized feature vector:
\begin{equation}
\bar{f}_y = \frac{1}{N_y} \sum_{i=1}^{N_y}{f}_i, \hat{f}_y = \frac{\bar{f}_y}{\| \bar{f}_y \|_2}.
\end{equation}
The semantic reward is then defined by the cosine similarity between the generated feature and the class prototype:
\begin{equation}
R_{\text{sem}} = \frac{1}{2} 
\left( \hat{f}_g^\top \hat{f}_y + 1 \right),
\end{equation}
where $\hat{f}_g$ denotes the normalized feature of the generated image.
This similarity is linearly rescaled to $[0,1]$, encouraging the generator to produce identity-consistent representations in the target feature space.

\subsubsection{Distributional Coverage}
While semantic consistency enforces identity preservation, it may constrain the generator to a limited region of the feature space. To encourage exploration of a broader range of intra-class variations, we introduce a kernel-based coverage reward that compares the distribution of generated features with the corresponding reference distribution in the memory bank.
Let $\hat{\mathcal{G}}_y = \{ \hat{f}_{g,j} \}_{j=1}^{B}$ denote the normalized features of the generated samples of current batch with batchsize $B$, and $\hat{\mathcal{B}}_y=\{\hat{f}_{y,i}\}_{i=1}^{N_y}$ denote the normalized reference features.
We use a radial basis function (RBF) kernel that measures pairwise similarity in the feature space~\cite{li2015generative}, with bandwidth $\sigma$ controlling sensitivity to local variations,
\begin{equation}
k_\sigma(u,v) = \exp\!\left(-{\|u - v\|_2^2}/{2\sigma^2}\right),
\end{equation}
and define the diversity reward as
\begin{equation}
\footnotesize
R_{\text{cov}} =
\mathbb{E}_{g\in{\hat{\mathcal{G}}_y},r\in{\hat{\mathcal{B}}_y}}\!\left[ k_\sigma(\hat{f}_g, \hat{f}_r) \right]
- \alpha \, \mathbb{E}_{g,g'\in{\hat{\mathcal{G}}_y}}\!\left[ k_\sigma(\hat{f}_g, \hat{f}_{g'}) \right],
\end{equation}
where the coefficient $\alpha>0$ controls the trade-off between distributional alignment and redundancy suppression.
The first term encourages distributional alignment between generated and reference features, whereas the second term penalizes redundancy among generated samples. This formulation promotes intra-class coverage while mitigating mode collapse, fostering a balanced trade-off between representational alignment and diversity.

\subsubsection{Expressive Diversity}
While the previous part promotes intra-class coverage, it does not explicitly control the overall spread of the generated feature distribution. To regulate the global dispersion of generated features and prevent over-concentration or under-dispersion, we define a covariance expansion reward based on the trace of the feature covariance matrices,
\begin{align}
\Sigma_g &= \frac{1}{B-1} \sum_{j=1}^{B} 
(\hat{f}_{g,j} - \bar{f}_g)
(\hat{f}_{g,j} - \bar{f}_g)^\top, \\
\Sigma_r &= \frac{1}{N_y-1} \sum_{i=1}^{N_y} 
(\hat{f}_{y,i} - \bar{f}_y)
(\hat{f}_{y,i} - \bar{f}_y)^\top,
\end{align}
where $\bar{f}_g$ and $\bar{f}_y$ denote the corresponding feature means.  
We use their traces
\begin{equation}
S_g = \operatorname{tr}(\Sigma_g), \quad 
S_r = \operatorname{tr}(\Sigma_r),
\end{equation}
to characterize the overall feature variance.  
The target variance is set as $(1+\varepsilon) S_r$,  
and the covariance expansion reward is formulated as,
\begin{equation}
R_{\text{exp}} = 
- \left({S_g - (1+\varepsilon) S_r}/{\tau} \right)^2,
\end{equation}
which softly encourages the generated feature distribution to maintain a controlled level $\varepsilon$ of expansion relative to the reference distribution.

To ensure numerical stability and comparability across heterogeneous terms,
each reward component is standardized by its batch-wise mean and standard deviation:
\begin{equation}
\tilde{R}_i = 
\frac{R_i - \mu_i}{\sigma_i + \epsilon}, 
\quad i \in \{ \text{sem}, \text{cov}, \text{exp} \}.
\end{equation}
The final normalized total reward is
\begin{equation}
R_{\text{norm}} = 
\tanh\!\left(
\lambda_{\text{sem}} \tilde{R}_{\text{sem}} +
\lambda_{\text{cov}} \tilde{R}_{\text{cov}} +
\lambda_{\text{exp}} \tilde{R}_{\text{exp}}
\right),
\end{equation}
where $\lambda_\text{sem}$, $\lambda_\text{cov}$, $\lambda_\text{exp}$ control the relative importance of each component. $\tanh(\cdot)$ activation bounds the overall reward to a stable numerical range for optimization.

\subsection{Dynamic Sample Selection Strategy}
After RL fine-tuning, distributional discrepancies between synthesized and real data may persist, resulting in uneven training utility across synthetic samples. To address this, we propose a lookahead-guided strategy that dynamically selects high-utility synthetic samples during 
training.

At each iteration, a mixed batch containing both real and synthetic samples is constructed across multiple identities. A one-step virtual gradient update is simulated on this batch to approximate the current optimization direction. The utility of each candidate synthetic sample $\hat{x}$ is then estimated as the change in identity-specific loss,
\begin{equation}
\Delta l = l_{\text{id}}(\bm{w}', \hat{\bm{x}}) - l_{\text{id}}(\bm{w}, \hat{\bm{x}}),
\end{equation}
where $\bm{w}$ and $\bm{w}'$ denote the model parameters before and after the virtual update, respectively, and $l_\text{id}(\cdot)$ is the identity-consistency loss. 
A smaller $\Delta l$ indicates that the sample better aligns with the ongoing optimization trajectory.

Synthetic samples with the smallest $\Delta l$ are selected to form a refined batch for model updates. This lookahead-guided mechanism ensures that gradient steps are influenced by synthetic samples most compatible with the current model state, leading to more stable training and improved generalization under distributional shifts.

\definecolor{myred}{RGB}{210,30,10} 
\definecolor{myblue}{RGB}{0,100,170} 
\definecolor{mygreen}{RGB}{20,155,90} 
\definecolor{light-blue}{RGB}{230, 245, 255}
\definecolor{light-gray}{gray}{0.9}

\begin{table}
\centering
\begin{small}
\setlength{\tabcolsep}{1.2pt}
\caption{Comparisons with different synthesis-based SOTA methods on {Market-1501} and CUHK03-NP. The comparison results are reproduced in our implementation to ensure fair and consistent evaluation. mAP(\%) and Rank-1 (\%) accuracy are reported.} 
\vspace{-2mm}
\begin{tabular}{clcccc}
\toprule
\multirow{2}{*}{Types} & \multirow{2}{*}{Methods} & \multicolumn{2}{c}{Market-1501} & \multicolumn{2}{c}{CUHK03} \\
\cmidrule{3-4}\cmidrule{5-6}
& & mAP & rank-1 & mAP & rank-1 \\
\midrule
Base & ResNet-50~\cite{cvpr/HeZRS16}$^{\dag}$ & $85.4$ & $85.4$ & $74.1$ & $76.5$ \\
\midrule
\multirow{3}{*}{Real Aug.} 
& R-Erasing~\cite{aaai/Zhong0KL020}$^{\dag}$\tiny{AAAI'20} & $87.6 $& $94.8$ & $76.7$ & $78.4$ \\
& CIDAM~\cite{hong2022camera} \tiny{ACMMM'22} & $87.4$ & $95.1$ & - & - \\
& CaAug~\cite{liu2024cloth}$^{\dag}$\tiny{ACMMM'24} & $86.4$ & $94.4$ & $57.4$ & $59.4$ \\
\midrule
\multirow{3}{*}{\shortstack{Simulated\\Aug.}} 
& FineGPR~\cite{xiang2023less} \tiny{TOMM'23} & $82.4$ & $92.6$ & $36.4$ & $37.9$ \\
& InfnitePerson~\cite{zhang2024infiniteperson} \tiny{TCSVT'25} & $57.3$ & $79.6$ & $24.7$ & $24.6$ \\
& VIPerson~\cite{zhang2025viperson}~\tiny{ICCV'25} & $86.9$ & $95.1$ & - & - \\
\midrule
\multirow{4}{*}{\shortstack{Synthetic\\Aug.}} 
& DG-Net~\cite{cvpr/ZhengYY00K19} \tiny{CVPR'19} & $86.0$ & $94.8 $& - & - \\
& GIF-SD~\cite{zhang2023expanding}$^{\dag}$\tiny{NeurIPS'23} & $74.9$ & $88.9$ & $71.7$ & $74.6$ \\
& IDiff~\cite{boutros2023idiff}$^{\dag}$\tiny{CVPR'23} & $85.4$ & $94.4$ & $73.1$ & $75.4$ \\
\rowcolor{light-blue}
& Ours & 8$8.6$ & 9$4.9$ & $76.6$ & $79.3$ \\
\bottomrule
\end{tabular}
\label{tab:reid_results}
\end{small}
\end{table}

\begin{table*}
\centering
\begin{small}
\setlength{\tabcolsep}{8pt}
\caption{Comparison of on the proposed method with SOTA trained on small-scale CASIA-WebFace~\cite{yi2014learning} subset. The highest and second-highest verification accuracies (\%) are highlighted in \textcolor{myred}{red} and \textcolor{myblue}{blue}, respectively. }
\begin{tabular}{clcccccc}
\toprule
{\textbf{Data Generation}} & {\textbf{Methods}} & {\textbf{LFW}} & {\textbf{AgeDB}} & \textbf{CFP-FP} & {\textbf{CA-LFW}} & {\textbf{CP-LWF}} & {\textbf{Avg.}} \\
\midrule
Authentic & CASIA-Webface~\cite{yi2014learning} subset & $91.58\%$ & $74.72\%$ &$ 76.00\%$ & $78.78\%$ & $71.15\%$ & $78.47\%$ \\
\midrule
\multirow{4}{*}{GAN} & SFace~\cite{boutros2022sface} \tiny{IJCB'22} & $85.40\%$ & $66.82\%$ & $69.14\%$ & $71.50\%$ & $67.35\%$ & $72.04\%$ \\
& IDnet~\cite{kolf2023identity} \tiny{CVPR'23} & $85.53\%$ & $68.73\%$ & $69.91\%$ & $72.67\%$ & $68.12\%$ & $73.00\%$ \\
& SFace2~\cite{boutros2024sface2} \tiny{T-BIOM'24} & $85.58\%$ & $68.12\%$ & $69.26\%$ & $72.35\%$ & $66.83\%$ & $72.43\%$ \\
& DCFace~\cite{kim2023dcface} \tiny{CVPR'23} & $87.97\%$ & $69.75\%$ & $66.33\%$ & $76.53\%$ & $64.05\%$ & $72.96\%$ \\
\midrule
\multirow{3}{*}{DM} & IDiff-Face~\cite{boutros2023idiff} \tiny{CVPR'23} & $90.65\%$ & $66.60\%$ & \textcolor{myred}{$75.64\%$} & $75.42\%$ & $68.70\%$ & 7$5.40\%$ \\
& NegFaceDiff~\cite{caldeira2025negfacediff} \tiny{CVPR'25} & $91.70\%$ & $74.68\%$ & \textcolor{myblue}{$75.06\%$} & $78.67\%$ & \textcolor{myred}{$70.53\%$} & \textcolor{myblue}{$78.13\%$} \\
\rowcolor{light-blue}
& Ours & \textcolor{myred}{$93.60\%$} & \textcolor{myred}{$76.80\%$} & $73.26\%$ & \textcolor{myred}{$81.68\%$} & \textcolor{myblue}{$70.02\%$} & \textcolor{myred}{$79.07\%$} \\
\bottomrule
\end{tabular}
\label{tab:face_results}
\end{small}
\end{table*}

\begin{table}
\centering
\small
\setlength{\tabcolsep}{3.2pt}
\caption{Demographic bias assessment of face recognition models trained with our method and SOTA approaches. The ethnicity-specific results report verification accuracies (\%) on each subset of RFW~\cite{wang2019racial}. }
\vspace{-2mm}
\begin{tabular}{lccccc} 
\toprule
\textbf{Methods} & \textbf{Caucasian} & \textbf{Indian} & \textbf{Asian} & \textbf{African} & \textbf{Avg.} \\
\midrule
Authentic~\cite{yi2014learning} & $72.85$ & {$70.10$} & $65.45$ & $60.28$ & $67.17$ \\
\midrule
SFace~\cite{boutros2022sface} & $67.17$ & $66.07$ & $62.92$ & $56.17$ & $63.08$ \\
IDnet~\cite{kolf2023identity} & $69.27$ & $66.83$ & $64.77$ & $57.85$ & $64.68$ \\
DCFace~\cite{kim2023dcface} & $69.80$ & $65.82$ & \textcolor{myred}{$69.80$} & $57.97$ & $65.85$ \\
SFace2~\cite{boutros2024sface2} & $67.78$ & $66.28$ & $63.68$ & $58.22$ & $63.99$ \\
IDiff-Face~\cite{boutros2023idiff} & $70.78$ & $67.10$ & $66.18$ & $58.65$ & $65.68$ \\
NegFaceDiff~\cite{caldeira2025negfacediff} & \textcolor{myblue}{$72.15$} & \textcolor{myblue}{$69.78$} & $67.07$ & \textcolor{myblue}{$60.13$} & \textcolor{myblue}{$67.28$}\\
\rowcolor{light-blue}
Ours & \textcolor{myred}{$75.87$} & \textcolor{myred}{$71.85$} & \textcolor{myblue}{$68.72$} & \textcolor{myred}{$62.67$} & \textcolor{myred}{$69.78$} \\
\bottomrule
\end{tabular}
\label{tab:ethnic}
\end{table}

\section{Experiments}
\subsection{Experimental Setup}
\noindent\textbf{Datasets and Evaluation Protocols.}
Focusing on task domains where real data collection is inherently constrained by privacy restrictions, we evaluate our approach on two identity-related tasks: person re-identification and face recognition.
To enable fair evaluation and comparison, we employ two small-scale person re-identification datasets, Market-1501~\cite{iccv/ZhengSTWWT15} and CUHK03-NP~\cite{cvpr/LiZXW14} datasets, along with a subset of the CASIA dataset~\cite{corr/YiLLL14a} for face recognition.
{LFW}~\cite{huang2008labeled}, {AgeDB}~\cite{cvpr/MoschoglouPSDKZ17}, {CFP-FP}~\cite{wacv/SenguptaCCPCJ16}, CA-LFW~\cite{zheng2017cross}, and CP-LFW~\cite{zheng2018cross} are used for downstream face verification, RFW~\cite{wang2019racial} is used for demographic bias assessment.

\noindent\textbf{Implementing Details.} All the experiments are implemented with NVIDIA H200 GPUs using pytorch. 
\textit{Cold-start.} We adopt DiT-XL/2~\cite{iccv/PeeblesX23} pretrained on ImageNet~\cite{cvpr/DengDSLL009} as the base backbone, and set learning rate as $1e-5$, the person image size at $256 \times256$, and the face image size at $128 \times128$. We reset the weights of the class embedding layer to zero before training. The embedding was then learned from scratch using the target dataset labels to ensure compatibility with the new label space.
\textit{Reward Optimization.} DPOK algorithm~\cite{fan2023dpok}
guides policy gradient optimization learning rate of
$1e-5$. $\lambda_\text{sem}$, $\lambda_\text{cov}$, and $\lambda_\text{exp}$ are at $1.0$, $0.75$, and $0.25$.
\textit{Downstreaming Task.}
For re-identification training, we use
ResNet-50~\cite{cvpr/HeZRS16} and ViT-16~\cite{iclr/DosovitskiyB0WZ21} as the backbone with Adam optimizer~\cite{iclr/KingmaB14}, setting the weight decay to $5e-4$. The input images are resized to $256 \times 128$ for training. The initial mini-batch size is set to 64, containing $P=16$ persons with $N=4$ images each. The loss function is a combination of ID classification loss (cross-entropy) and metric learning loss (triplet loss with hard mining~\cite{schroff2015facenet}). For face recognition training, we use ResNet-50~\cite{cvpr/HeZRS16} with CosFace~\cite{wang2018cosface}.
We set the mini-batch size to $128$ and train with SGD optimizer~\cite{robbins1951stochastic}, setting the momentum to $0.9$ and the weight decay to $5e-4$.

\begin{figure*}[t]
    \centering
    \includegraphics[width=0.8\textwidth]{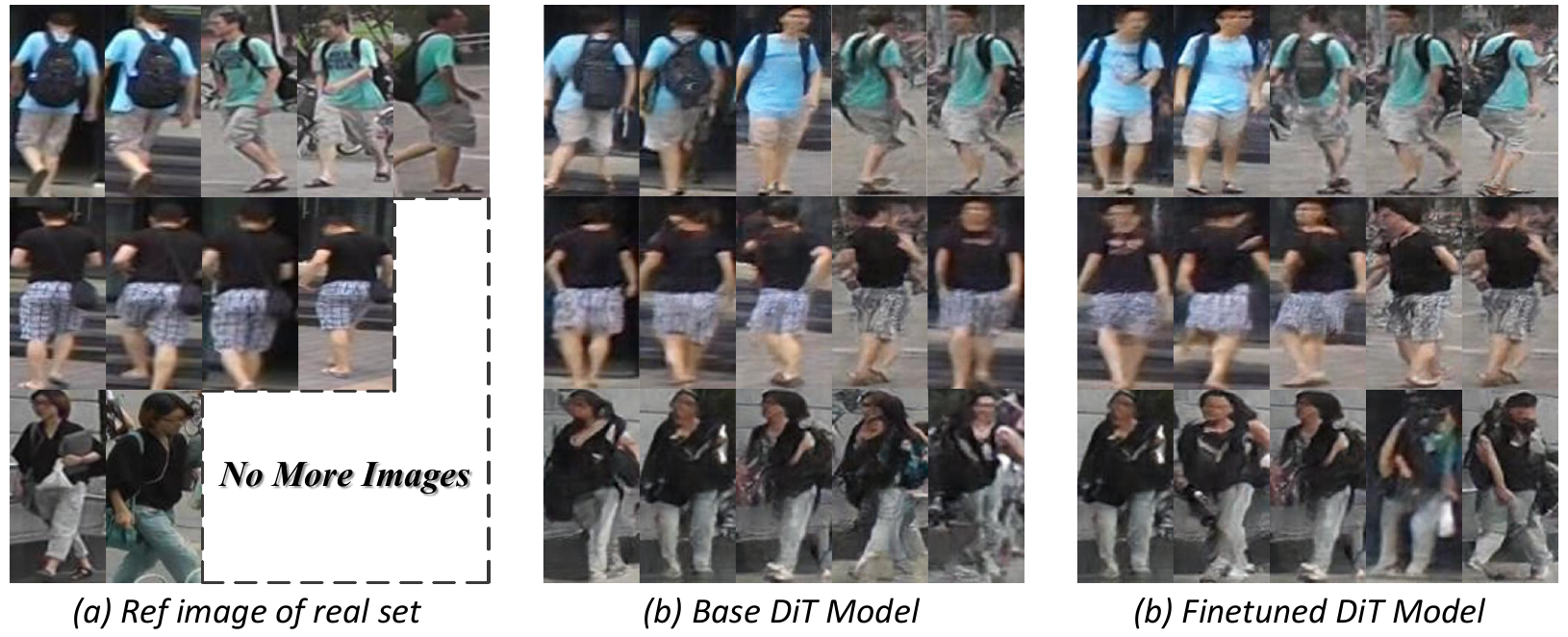} 
    \vspace{-2mm}
    \caption{Comparisons with the baseline on Market-1501 generation. Real reference images are randomly selected from training set, where certain identity classes have only a few samples. While the baseline DiT benefits from external ImageNet pretraining to introduce moderate diversity, our RL-based fine-tuning further enhances intra-class variability, generating more diverse yet identity-consistent images.} 
    \label{market_vis}
    \vspace{-2mm}
\end{figure*}

\begin{figure*}[t]
    \centering
    \includegraphics[width=0.8\textwidth]{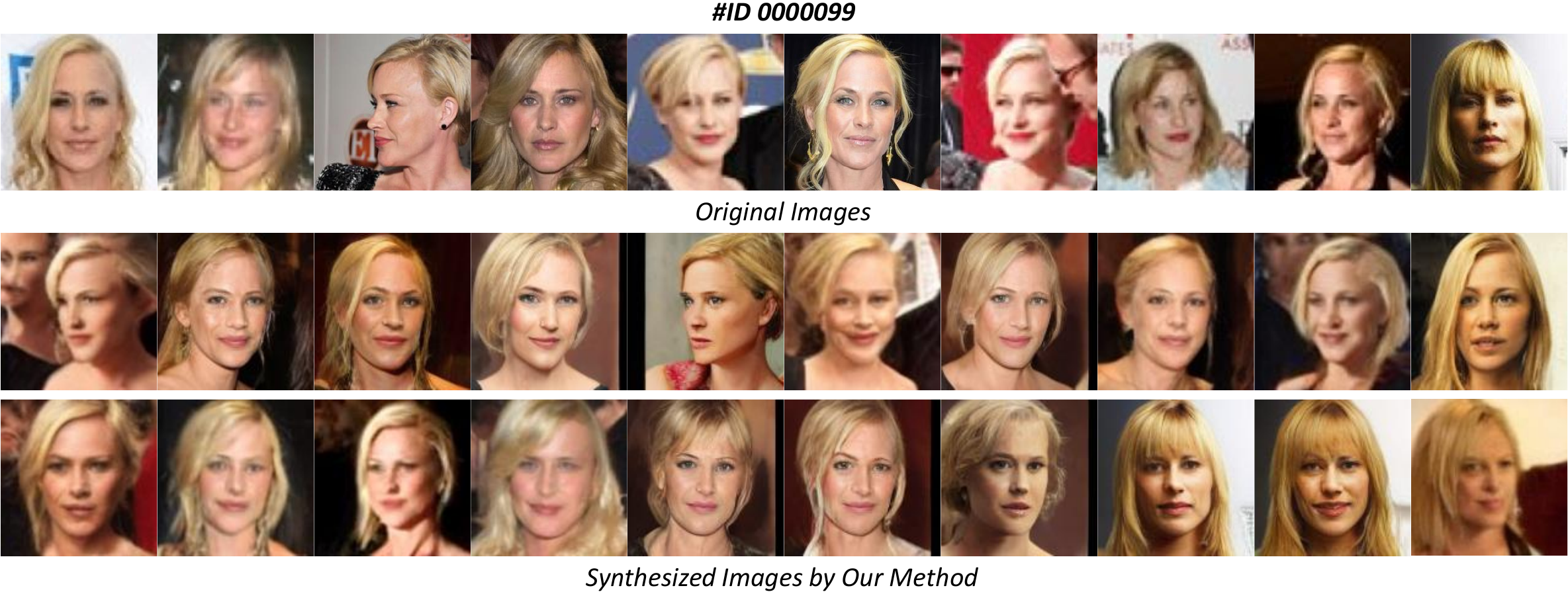} 
    \caption{Samples generated by our method. Real images are randomly selected from the training set of CASIA-WebFace, where certain identity classes have only a few available samples. While the baseline DiT benefits from external ImageNet pretraining to introduce moderate diversity, our RL-based fine-tuning generates more diverse images while largely preserving identity characteristics. }
    \label{face_vis}
    \vspace{-2mm}
\end{figure*}

\subsection{Quantitative Results.}
\noindent\textbf{Person Re-identification Evaluation} We categorize mainstream methods into three groups: real-image augmentation~\cite{aaai/Zhong0KL020,hong2022camera,liu2024cloth}, simulation-based expansion~\cite{xiang2023less,zhang2024infiniteperson,zhang2025viperson}, and synthetic enhancement. For the baseline, we apply standard data augmentation techniques, including horizontal flipping and padding.
Table~\ref{tab:reid_results} compares three categories of approaches on Market-1501 and CUHK03-NP.
Overall, real-image augmentation methods provide stable but limited improvements over the ResNet-50 baseline, yielding about a $2\%$ mAP gain on Market-1501. Such pixel-level perturbations slightly increase appearance variation but fail to introduce new identity structures.
Simulation-based methods perform worse due to the large domain gap between virtual and real imagery. Without costly environment calibration or domain alignment, their mAP stays low, far behind real-data augmentations.
Among generation-based approaches, performance varies significantly.
GIF-SD~\cite{zhang2023expanding}, which generates images for general classification via text-guided synthesis, performs poorly in identity recognition since highly similar identity appearances limit the effectiveness of textual guidance. Consequently, its generated samples deviate from real distributions and even degrade training (mAP $\downarrow$ $10.5\%$ \textit{vs.} baseline).
In contrast, our reinforcement-guided method adapts general-domain priors to the target domain through a task-specific reward, achieving $88.6\%$ mAP ($3.2\%$ gain) on Market-1501 and $76.6\%$ ($2.5\%$ gain) on CUHK03. This demonstrates that reward-driven synthesis generates high-fidelity, diverse, and identity-preserving samples that consistently improve downstream recognition.

\noindent\textbf{Face Recognition Evaluation.} Table~\ref{tab:face_results} reports verification results on the small-scale CASIA-WebFace~\cite{yi2014learning} subset.
GAN-based methods show limited improvement, with average accuracies remaining around $72\sim73\%$, indicating that their generated samples provide weak identity supervision under scarce data.
Diffusion-based models (DM) achieve higher fidelity, yet previous approaches such as IDiff-Face~\cite{boutros2023idiff} and NegFaceDiff~\cite{caldeira2025negfacediff} still fall short in maintaining consistent identity cues.
In contrast, our reinforcement-guided model achieves the best overall performance with an average accuracy of $79.07\%$, surpassing NegFaceDiff~\cite{caldeira2025negfacediff} by $0.94\%$ and the real-data baseline by $0.60\%$.
These results confirm that our reward-driven adaptation enhances both generation fidelity and task utility, especially in data-limited face recognition scenarios.

\noindent\textbf{Racial Bias Assessment.}
Table~\ref{tab:ethnic} evaluates model bias across ethnic subsets of RFW.
Our method achieves the highest average accuracy of $69.78\%$, surpassing all competing approaches.
Compared with IDiff-Face~\cite{boutros2023idiff} and NegFaceDiff~\cite{caldeira2025negfacediff}, our model shows more balanced performance across groups, especially improving on the Indian ($1.0\%$) and Asian ($0.9\%$) subsets.
These results demonstrate that the proposed reinforcement-guided synthesis not only enhances overall accuracy but also mitigates cross-ethnicity bias in face recognition.

\begin{figure}[t]
    \centering
    \includegraphics[width=\columnwidth]{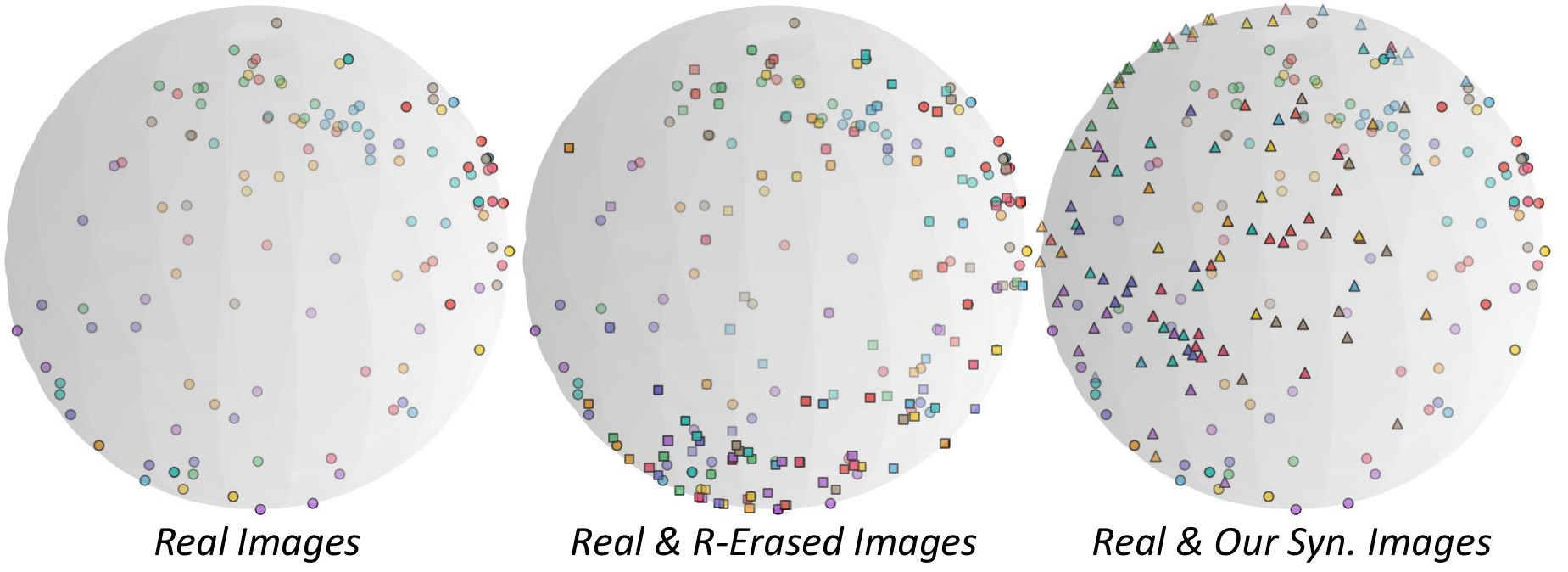} 
    \caption{Feature distributions of real and synthesized samples across different methods. Image embeddings are extracted using a pretrained ResNet-50 and projected into a shared space via DOSNES~\cite{lu2016doubly}. Circles, squares, and triangles denote real samples, Random-Erasing synthesized samples, and synthesized samples generated by our method, respectively. Each color represents an identity class, with ten randomly selected classes visualized.}
    \label{tsne}
\end{figure}

\subsubsection{Qualitative Results}

As illustrated in Figure~\ref{market_vis}, our reinforcement-guided fine-tuning substantially improves both the quality and diversity of generated pedestrian images on the Market-1501 dataset.
The Base DiT model, although benefiting from external ImageNet pretraining, produces limited intra-class variation and often replicates similar appearances, especially when the original identity class contains only a few real samples.
In contrast, the finetuned DiT model guided by our reinforcement strategy generates richer pose, viewpoint, and illumination variations while maintaining consistent identity cues.
Even under data-scarce conditions, our method can synthesize high-quality and diverse samples, demonstrating its strong potential for applications in privacy-sensitive or limited-data scenarios.

As shown in Figure~\ref{face_vis}, our method generates a wide range of face images that retain the subject’s overall identity while presenting clear variations in expression, pose, hairstyle, and illumination.
Compared with the limited original samples in CASIA-WebFace, the synthesized results exhibit richer intra-class diversity and higher visual realism, effectively compensating for the scarcity of real data.
These results demonstrate that the proposed reinforcement-guided synthesis can extend the visual representation of each identity, providing a more comprehensive and diverse dataset foundation for recognition tasks.

In Figure~\ref{tsne}, the feature visualization reveals that our synthesized samples occupy a broader and more continuous region in the embedding space compared with baseline augmentations.
In the DOSNES~\cite{lu2016doubly} projection, real samples form compact clusters, while those generated by Random-Erasing remain close to the original data and show limited variation.
By contrast, the features of our synthesized images expand the local neighborhood of each identity cluster, enriching intra-class structures without drifting away from the original centers.
This demonstrates that our reinforcement-guided generation introduces meaningful diversity while preserving identity consistency, resulting in a more complete and balanced representation manifold.

\subsection{Ablation Studies}

\begin{figure}[t]
    \centering
    \includegraphics[width=\columnwidth]{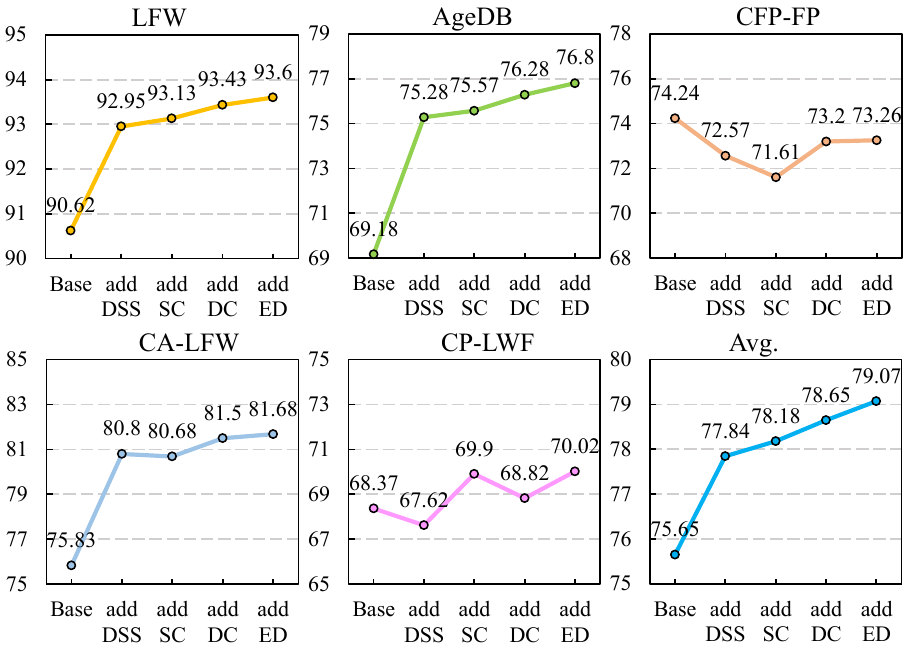} 
    \caption{Ablation studies of our proposed method. Adding components consistently improve the face vertification accuracies (\%). } 
    \label{fig:ablation}
\end{figure}

\textbf{Components Effectiveness.}
As shown in Figure~\ref{fig:ablation}, each proposed component consistently improves the baseline performance.
Starting from Base-DiT ($75.65\%$), introducing Dynamic Sample Selection (DSS) yields a clear gain of $2.2\%$, demonstrating that adaptively emphasizing task-relevant samples enhances data utility.
Further integration of semantic reward (SC) improves average accuracy to $78.18\%$, indicating that enforcing semantic consistency helps preserve identity-related information.
The coverage reward (DC) brings another improvement to $78.65\%$ by promoting feature coverage and intra-class diversity.
Finally, adding the expression reward (ED) achieves the best overall result of $79.07\%$, confirming that encouraging expression richness strengthens model robustness under appearance variations.
Overall, the progressive gains validate the complementary effects of the designed rewards and the effectiveness of the reinforcement-guided synthesis method.

\begin{figure}[t]
    \centering
    \includegraphics[width=\columnwidth]{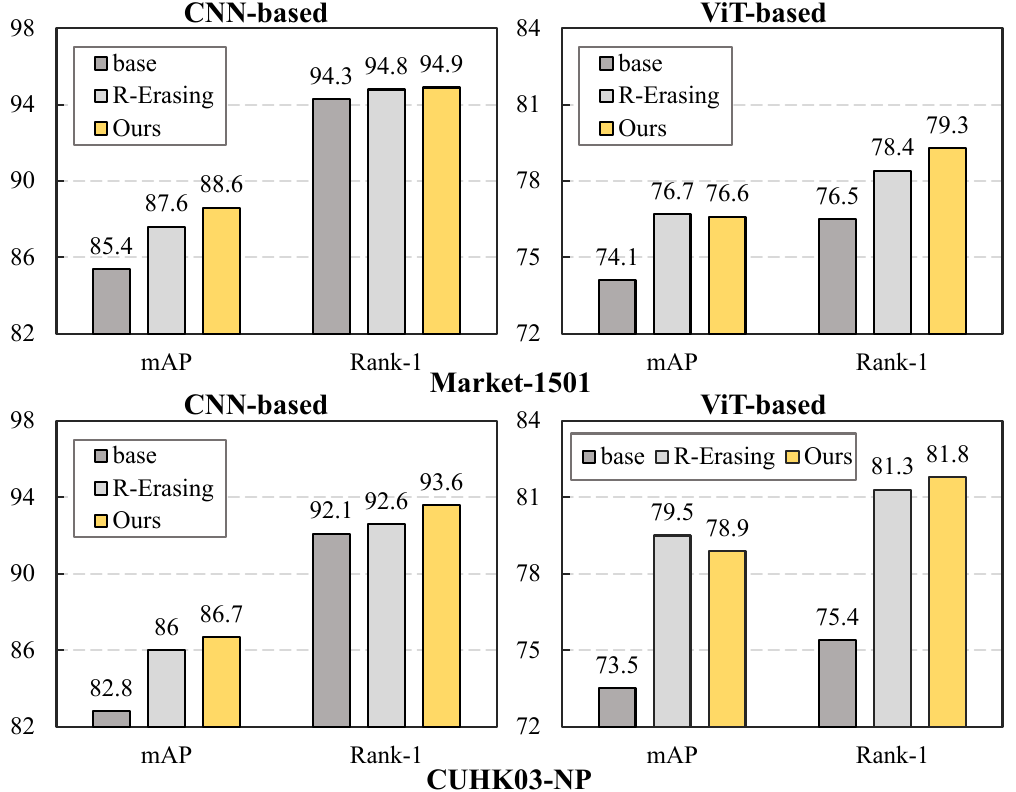} 
    \vspace{-3mm}
    \caption{Comparison of baseline models with and without our method across person ReID datasets. }
    \label{fig:baseline}
\end{figure}

\noindent\textbf{Downstream Backbone Effectiveness.}
As depicted in Figure~\ref{fig:baseline}, our method improves discriminative cues across backbones with minimal tuning. On CNN, it delivers balanced gains, exceeding Base by 0.6 Rank-1 and 3.9 mAP on the two datasets. On ViT, it yields clear Rank-1 boosts while keeping mAP comparable to R-Erasing, indicating robust compatibility with attention-based backbones.

\noindent\textbf{Discussion and Limitation}
While our proposed framework effectively enhances sample diversity and task relevance under limited data conditions, several limitations remain. The method still relies on the representational quality of the pretrained backbone, which may restrict performance in domains with poor prior alignment. In addition, the reward function requires task-specific tuning to balance utility and diversity. Our current study focuses on image-based privacy-sensitive tasks; extending to other modalities such as video or event data is left for future work. Despite these limitations, we believe the proposed approach offers a promising step toward data-efficient, privacy-aware generation guided by general-domain priors.

\section{Conclusion}
In this work, we present a reinforcement-guided framework for synthetic data generation tailored to privacy-sensitive identity recognition tasks.
By leveraging general-domain priors and casting synthesis as a reinforcement learning problem, the method enables task-oriented generation under limited data.
A multi-objective reward balancing semantic consistency, diversity, and expression richness guides the generator to produce visually realistic and utility-effective samples for downstream recognition.
In addition, a dynamic sample selection strategy further improves generalization by emphasizing task-relevant samples during training.
Experiments on face recognition and person re-identification benchmarks show that the proposed approach effectively alleviates data scarcity, achieving superior performance and robustness compared with existing distribution-matching methods.
We believe that this study provides a promising step toward privacy-aware and utility-driven data synthesis, paving the way for broader applications of generative models in identity-sensitive domains. 

\section*{Acknowledgements.} This work was supported by the National Natural Science Foundation of China (62571379) and the National Research Foundation, Singapore and Infocomm Media Development Authority under its Trust Tech Funding Initiative. Any opinions, findings and conclusions or recommendations expressed in this material are those of the author(s) and do not reflect the views of the National Research Foundation, Singapore and Infocomm Media Development Authority.
{
    \small
    \bibliographystyle{ieeenat_fullname}
    \bibliography{main}

@String(AAAI = {AAAI})

@article{tcsvt/ZhangLZW24,
  author={Guoqing Zhang and 
          Jin Li and 
          Yuhui Zheng and 
          Ruili Wang},
  journal={IEEE Transactions on Circuits and Systems for Video Technology}, 
  title={InfinitePerson: Innovating Synthetic Data Creation for Generalization Person Re-Identification}, 
  year={2024},
  volume={},
  number={},
  pages={1-1}
}

@inproceedings{cvpr/RombachBLEO22,
  author       = {Robin Rombach and
                  Andreas Blattmann and
                  Dominik Lorenz and
                  Patrick Esser and
                  Bj{\"{o}}rn Ommer},
  title        = {High-Resolution Image Synthesis with Latent Diffusion Models},
  booktitle    = {{IEEE/CVF} Conference on Computer Vision and Pattern Recognition},
  pages        = {10674--10685},
  year         = {2022}
}

@article{tmm/LuoJGLLLG20,
  author       = {Hao Luo and
                  Wei Jiang and
                  Youzhi Gu and
                  Fuxu Liu and
                  Xingyu Liao and
                  Shenqi Lai and
                  Jianyang Gu},
  title        = {A Strong Baseline and Batch Normalization Neck for Deep Person Re-Identification},
  journal      = {{IEEE} Trans. Multim.},
  volume       = {22},
  number       = {10},
  pages        = {2597--2609},
  year         = {2020}
}

@inproceedings{cvpr/ZhengYY00K19,
  author       = {Zhedong Zheng and
                  Xiaodong Yang and
                  Zhiding Yu and
                  Liang Zheng and
                  Yi Yang and
                  Jan Kautz},
  title        = {Joint Discriminative and Generative Learning for Person Re-Identification},
  booktitle    = {{IEEE} Conference on Computer Vision and Pattern Recognition},
  pages        = {2138--2147},
  year         = {2019}
}

@inproceedings{nips/HoJA20,
  author       = {Jonathan Ho and
                  Ajay Jain and
                  Pieter Abbeel},
  title        = {Denoising Diffusion Probabilistic Models},
  booktitle    = {Advances in Neural Information Processing Systems},
  year         = {2020}
}

@inproceedings{cvpr/Sun019,
  author       = {Xiaoxiao Sun and
                  Liang Zheng},
  title        = {Dissecting Person Re-Identification From the Viewpoint of Viewpoint},
  booktitle    = {{IEEE} Conference on Computer Vision and Pattern Recognition},
  pages        = {608--617},
  year         = {2019}
}

@misc{corr/abs-2211-11165,
  author       = {Likai Wang and
                  Xiangqun Zhang and
                  Ruize Han and
                  Jialin Yang and
                  Xiaoyu Li and
                  Wei Feng and
                  Song Wang},
  title        = {A Benchmark of Video-Based Clothes-Changing Person Re-Identification},
  archivePrefix      = {arXiv},
  eprint       = {2211.11165},
  year         = {2022}
}

@misc{corr/abs-2406-06045,
  author       = {Ke Niu and
                  Haiyang Yu and
                  Xuelin Qian and
                  Teng Fu and
                  Bin Li and
                  Xiangyang Xue},
  title        = {Synthesizing Efficient Data with Diffusion Models for Person Re-Identification
                  Pre-Training},
  archivePrefix      = {arXiv},
  eprint       = {2406.06045},
  year         = {2024},
}

@article{tip/JiaZYLH22,
  author       = {Xuemei Jia and
                  Xian Zhong and
                  Mang Ye and
                  Wenxuan Liu and
                  Wenxin Huang},
  title        = {Complementary Data Augmentation for Cloth-Changing Person Re-Identification},
  journal      = {{IEEE} Trans. Image Process.},
  volume       = {31},
  pages        = {4227--4239},
  year         = {2022}
}

@inproceedings{aaai/Zhong0KL020,
  author       = {Zhun Zhong and
                  Liang Zheng and
                  Guoliang Kang and
                  Shaozi Li and
                  Yi Yang},
  title        = {Random Erasing Data Augmentation},
  booktitle    = {{AAAI} Conference on Artificial Intelligence},
  pages        = {13001--13008},
  year         = {2020}
}

@inproceedings{wacv/KansalWK24,
  author       = {Kajal Kansal and
                  Yongkang Wong and
                  Mohan S. Kankanhalli},
  title        = {Privacy-Enhancing Person Re-identification Framework - {A} Dual-Stage
                  Approach},
  booktitle    = {{IEEE/CVF} Winter Conference on Applications of Computer Vision},
  pages        = {8528--8537},
  year         = {2024}
}

@inproceedings{iccv/PeeblesX23,
  author       = {William Peebles and
                  Saining Xie},
  title        = {Scalable Diffusion Models with Transformers},
  booktitle    = {International Conference on Computer Vision},
  pages        = {4172--4182},
  year         = {2023}
}

@inproceedings{cvpr/HeZRS16,
  author       = {Kaiming He and
                  Xiangyu Zhang and
                  Shaoqing Ren and
                  Jian Sun},
  title        = {Deep Residual Learning for Image Recognition},
  booktitle    = {{IEEE} Conference on Computer Vision and Pattern Recognition},
  pages        = {770--778},
  year         = {2016}
}

@inproceedings{iclr/KingmaB14,
  author       = {Diederik P. Kingma and
                  Jimmy Ba},
  title        = {Adam: {A} Method for Stochastic Optimization},
  booktitle    = {International Conference on Learning Representations},
  year         = {2015}
}

@inproceedings{iclr/DosovitskiyB0WZ21,
  author       = {Alexey Dosovitskiy and
                  Lucas Beyer and
                  Alexander Kolesnikov and
                  Dirk Weissenborn and
                  Xiaohua Zhai and
                  Thomas Unterthiner and
                  Mostafa Dehghani and
                  Matthias Minderer and
                  Georg Heigold and
                  Sylvain Gelly and
                  Jakob Uszkoreit and
                  Neil Houlsby},
  title        = {An Image is Worth 16x16 Words: Transformers for Image Recognition
                  at Scale},
  booktitle    = {International Conference on Learning Representations},
  year         = {2021}
}

@inproceedings{icml/NicholD21,
  author       = {Alexander Quinn Nichol and
                  Prafulla Dhariwal},
  title        = {Improved Denoising Diffusion Probabilistic Models},
  booktitle    = { International Conference on Machine Learning},
  volume       = {139},
  pages        = {8162--8171},
  year         = {2021}
}

@article{song2020denoising,
  title={Denoising diffusion implicit models},
  author={Song, Jiaming and Meng, Chenlin and Ermon, Stefano},
  journal={arXiv preprint arXiv:2010.02502},
  year={2020}
}

@inproceedings{nips/GoodfellowPMXWOCB14,
  author       = {Ian J. Goodfellow and
                  Jean Pouget{-}Abadie and
                  Mehdi Mirza and
                  Bing Xu and
                  David Warde{-}Farley and
                  Sherjil Ozair and
                  Aaron C. Courville and
                  Yoshua Bengio},
  title        = {Generative Adversarial Nets},
  booktitle    = {Advances in Neural Information Processing Systems},
  pages        = {2672--2680},
  year         = {2014}
}

@inproceedings{cvpr/DengDSLL009,
  author       = {Jia Deng and
                  Wei Dong and
                  Richard Socher and
                  Li{-}Jia Li and
                  Kai Li and
                  Li Fei{-}Fei},
  title        = {ImageNet: {A} large-scale hierarchical image database},
  booktitle    = {Computer Vision and Pattern
                  Recognition},
  pages        = {248--255},
  year         = {2009}
}

@inproceedings{nips/SahariaCSLWDGLA22,
  author       = {Chitwan Saharia and
                  William Chan and
                  Saurabh Saxena and
                  Lala Li and
                  Jay Whang and
                  Emily L. Denton and
                  Seyed Kamyar Seyed Ghasemipour and
                  Raphael Gontijo Lopes and
                  Burcu Karagol Ayan and
                  Tim Salimans and
                  Jonathan Ho and
                  David J. Fleet and
                  Mohammad Norouzi},
  title        = {Photorealistic Text-to-Image Diffusion Models with Deep Language Understanding},
  booktitle    = {Advances in Neural Information Processing Systems},
  year         = {2022}
}

@inproceedings{iccv/ZhengSTWWT15,
  author       = {Liang Zheng and
                  Liyue Shen and
                  Lu Tian and
                  Shengjin Wang and
                  Jingdong Wang and
                  Qi Tian},
  title        = {Scalable Person Re-identification: {A} Benchmark},
  booktitle    = {{IEEE} International Conference on Computer Vision},
  pages        = {1116--1124},
  year         = {2015}
}

@inproceedings{cvpr/LiZXW14,
  author       = {Wei Li and
                  Rui Zhao and
                  Tong Xiao and
                  Xiaogang Wang},
  title        = {DeepReID: Deep Filter Pairing Neural Network for Person Re-identification},
  booktitle    = {Conference on Computer Vision and Pattern Recognition},
  pages        = {152--159},
  year         = {2014}
}

@inproceedings{cvpr/MoschoglouPSDKZ17,
  author       = {Stylianos Moschoglou and
                  Athanasios Papaioannou and
                  Christos Sagonas and
                  Jiankang Deng and
                  Irene Kotsia and
                  Stefanos Zafeiriou},
  title        = {AgeDB: The First Manually Collected, In-the-Wild Age Database},
  booktitle    = {{IEEE} Conference on Computer Vision and Pattern Recognition
                  Workshops},
  pages        = {1997--2005},
  year         = {2017}
}

@inproceedings{wacv/SenguptaCCPCJ16,
  author       = {Soumyadip Sengupta and
                  Jun{-}Cheng Chen and
                  Carlos Domingo Castillo and
                  Vishal M. Patel and
                  Rama Chellappa and
                  David W. Jacobs},
  title        = {Frontal to profile face verification in the wild},
  booktitle    = {{IEEE} Winter Conference on Applications of Computer Vision},
  pages        = {1--9},
  year         = {2016}
}

@inproceedings{huang2008labeled,
  title={Labeled faces in the wild: A database forstudying face recognition in unconstrained environments},
  author={Huang, Gary B and Mattar, Marwan and Berg, Tamara and Learned-Miller, Eric},
  booktitle={Workshop on faces in'Real-Life'Images: detection, alignment, and recognition},
  year={2008}
}

@misc{corr/YiLLL14a,
  author       = {Dong Yi and
                  Zhen Lei and
                  Shengcai Liao and
                  Stan Z. Li},
  title        = {Learning Face Representation from Scratch},
  eprint       = {1411.7923},
  archivePrefix= "arXiv",
  year         = {2014}
}

@article{voigt2017eu,
  title={The eu general data protection regulation (gdpr)},
  author={Voigt, Paul and Von dem Bussche, Axel},
  journal={A practical guide, 1st ed., Cham: Springer International Publishing},
  volume={10},
  number={3152676},
  pages={10--5555},
  year={2017},
  publisher={Springer}
}

@article{amershi2014power,
  title={Power to the people: The role of humans in interactive machine learning},
  author={Amershi, Saleema and Cakmak, Maya and Knox, William Bradley and Kulesza, Todd},
  journal={AI magazine},
  volume={35},
  number={4},
  pages={105--120},
  year={2014}
}

@article{fan2023dpok,
  title={Dpok: Reinforcement learning for fine-tuning text-to-image diffusion models},
  author={Fan, Ying and Watkins, Olivia and Du, Yuqing and Liu, Hao and Ryu, Moonkyung and Boutilier, Craig and Abbeel, Pieter and Ghavamzadeh, Mohammad and Lee, Kangwook and Lee, Kimin},
  journal={Advances in Neural Information Processing Systems},
  volume={36},
  pages={79858--79885},
  year={2023}
}

@inproceedings{boutros2023idiff,
  title={Idiff-face: Synthetic-based face recognition through fizzy identity-conditioned diffusion model},
  author={Boutros, Fadi and Grebe, Jonas Henry and Kuijper, Arjan and Damer, Naser},
  booktitle={Proceedings of the IEEE/CVF International Conference on Computer Vision},
  pages={19650--19661},
  year={2023}
}

@article{zhang2023expanding,
  title={Expanding small-scale datasets with guided imagination},
  author={Zhang, Yifan and Zhou, Daquan and Hooi, Bryan and Wang, Kai and Feng, Jiashi},
  journal={Advances in neural information processing systems},
  volume={36},
  pages={76558--76618},
  year={2023}
}

@inproceedings{han2024advancing,
  title={Advancing text-driven chest x-ray generation with policy-based reinforcement learning},
  author={Han, Woojung and Kim, Chanyoung and Ju, Dayun and Shim, Yumin and Hwang, Seong Jae},
  booktitle={International Conference on Medical Image Computing and Computer-Assisted Intervention},
  pages={56--66},
  year={2024},
  organization={Springer}
}

@article{karras2020training,
  title={Training generative adversarial networks with limited data},
  author={Karras, Tero and Aittala, Miika and Hellsten, Janne and Laine, Samuli and Lehtinen, Jaakko and Aila, Timo},
  journal={Advances in neural information processing systems},
  volume={33},
  pages={12104--12114},
  year={2020}
}

@inproceedings{zheng2019joint,
  title={Joint discriminative and generative learning for person re-identification},
  author={Zheng, Zhedong and Yang, Xiaodong and Yu, Zhiding and Zheng, Liang and Yang, Yi and Kautz, Jan},
  booktitle={proceedings of the IEEE/CVF conference on computer vision and pattern recognition},
  pages={2138--2147},
  year={2019}
}

@inproceedings{wallace2024diffusion,
  title={Diffusion model alignment using direct preference optimization},
  author={Wallace, Bram and Dang, Meihua and Rafailov, Rafael and Zhou, Linqi and Lou, Aaron and Purushwalkam, Senthil and Ermon, Stefano and Xiong, Caiming and Joty, Shafiq and Naik, Nikhil},
  booktitle={Proceedings of the IEEE/CVF Conference on Computer Vision and Pattern Recognition},
  pages={8228--8238},
  year={2024}
}

@article{mirza2014conditional,
  title={Conditional generative adversarial nets},
  author={Mirza, Mehdi and Osindero, Simon},
  journal={arXiv preprint arXiv:1411.1784},
  year={2014}
}

@inproceedings{zhang2023adding,
  title={Adding conditional control to text-to-image diffusion models},
  author={Zhang, Lvmin and Rao, Anyi and Agrawala, Maneesh},
  booktitle={Proceedings of the IEEE/CVF international conference on computer vision},
  pages={3836--3847},
  year={2023}
}

@inproceedings{wu2023uncovering,
  title={Uncovering the disentanglement capability in text-to-image diffusion models},
  author={Wu, Qiucheng and Liu, Yujian and Zhao, Handong and Kale, Ajinkya and Bui, Trung and Yu, Tong and Lin, Zhe and Zhang, Yang and Chang, Shiyu},
  booktitle={Proceedings of the IEEE/CVF conference on computer vision and pattern recognition},
  pages={1900--1910},
  year={2023}
}

@inproceedings{schroff2015facenet,
  title={Facenet: A unified embedding for face recognition and clustering},
  author={Schroff, Florian and Kalenichenko, Dmitry and Philbin, James},
  booktitle={Proceedings of the IEEE conference on computer vision and pattern recognition},
  pages={815--823},
  year={2015}
}

@article{lu2016doubly,
  title={Doubly stochastic neighbor embedding on spheres},
  author={Lu, Yao and Corander, Jukka and Yang, Zhirong},
  journal={arXiv preprint arXiv:1609.01977},
  year={2016}
}

@article{yi2014learning,
  title={Learning face representation from scratch},
  author={Yi, Dong and Lei, Zhen and Liao, Shengcai and Li, Stan Z},
  journal={arXiv preprint arXiv:1411.7923},
  year={2014}
}

@inproceedings{kolf2023identity,
  title={Identity-driven three-player generative adversarial network for synthetic-based face recognition},
  author={Kolf, Jan Niklas and Rieber, Tim and Elliesen, Jurek and Boutros, Fadi and Kuijper, Arjan and Damer, Naser},
  booktitle={Proceedings of the IEEE/CVF Conference on Computer Vision and Pattern Recognition},
  pages={806--816},
  year={2023}
}

@inproceedings{boutros2022sface,
  title={Sface: Privacy-friendly and accurate face recognition using synthetic data},
  author={Boutros, Fadi and Huber, Marco and Siebke, Patrick and Rieber, Tim and Damer, Naser},
  booktitle={2022 IEEE International Joint Conference on Biometrics (IJCB)},
  pages={1--11},
  year={2022},
  organization={IEEE}
}

@article{boutros2024sface2,
  title={Sface2: Synthetic-based face recognition with w-space identity-driven sampling},
  author={Boutros, Fadi and Huber, Marco and Luu, Anh Thi and Siebke, Patrick and Damer, Naser},
  journal={IEEE Transactions on Biometrics, Behavior, and Identity Science},
  volume={6},
  number={3},
  pages={290--303},
  year={2024},
  publisher={IEEE}
}

@inproceedings{kim2023dcface,
  title={Dcface: Synthetic face generation with dual condition diffusion model},
  author={Kim, Minchul and Liu, Feng and Jain, Anil and Liu, Xiaoming},
  booktitle={Proceedings of the ieee/cvf conference on computer vision and pattern recognition},
  pages={12715--12725},
  year={2023}
}

@inproceedings{miao2024training,
  title={Training diffusion models towards diverse image generation with reinforcement learning},
  author={Miao, Zichen and Wang, Jiang and Wang, Ze and Yang, Zhengyuan and Wang, Lijuan and Qiu, Qiang and Liu, Zicheng},
  booktitle={Proceedings of the IEEE/CVF Conference on Computer Vision and Pattern Recognition},
  pages={10844--10853},
  year={2024}
}

@article{caldeira2025negfacediff,
  title={NegFaceDiff: The Power of Negative Context in Identity-Conditioned Diffusion for Synthetic Face Generation},
  author={Caldeira, Eduarda and Damer, Naser and Boutros, Fadi},
  journal={arXiv preprint arXiv:2508.09661},
  year={2025}
}

@inproceedings{saremi2025rl4med,
  title={RL4Med-DDPO: reinforcement learning for controlled guidance towards diverse medical image generation using vision-language foundation models},
  author={Saremi, Parham and Kumar, Amar and Mohamed, Mohamed and TehraniNasab, Zahra and Arbel, Tal},
  booktitle={International Conference on Medical Image Computing and Computer-Assisted Intervention},
  pages={478--488},
  year={2025},
  organization={Springer}
}

@inproceedings{wang2018cosface,
  title={Cosface: Large margin cosine loss for deep face recognition},
  author={Wang, Hao and Wang, Yitong and Zhou, Zheng and Ji, Xing and Gong, Dihong and Zhou, Jingchao and Li, Zhifeng and Liu, Wei},
  booktitle={Proceedings of the IEEE conference on computer vision and pattern recognition},
  pages={5265--5274},
  year={2018}
}

@inproceedings{wang2019racial,
  title={Racial faces in the wild: Reducing racial bias by information maximization adaptation network},
  author={Wang, Mei and Deng, Weihong and Hu, Jiani and Tao, Xunqiang and Huang, Yaohai},
  booktitle={Proceedings of the ieee/cvf international conference on computer vision},
  pages={692--702},
  year={2019}
}

@article{xiang2023less,
  title={Less is more: Learning from synthetic data with fine-grained attributes for person re-identification},
  author={Xiang, Suncheng and Qian, Dahong and Guan, Mengyuan and Yan, Binjie and Liu, Ting and Fu, Yuzhuo and You, Guanjie},
  journal={ACM Transactions on Multimedia Computing, Communications and Applications},
  volume={19},
  number={5s},
  pages={1--20},
  year={2023},
  publisher={ACM New York, NY}
}

@article{zhang2024infiniteperson,
  title={InfinitePerson: Innovating Synthetic Data Creation for Generalization Person Re-Identification},
  author={Zhang, Guoqing and Li, Jin and Zheng, Yuhui and Wang, Ruili},
  journal={IEEE Transactions on Circuits and Systems for Video Technology},
  year={2024},
  publisher={IEEE}
}

@inproceedings{hong2022camera,
  title={Camera-specific informative data augmentation module for unbalanced person re-identification},
  author={Hong, Pingting and Wu, Dayan and Li, Bo and Wang, Weipinng},
  booktitle={Proceedings of the 30th ACM International Conference on Multimedia},
  pages={501--510},
  year={2022}
}

@inproceedings{liu2024cloth,
  title={Cloth-aware Augmentation for Cloth-generalized Person Re-identification},
  author={Liu, Fangyi and Ye, Mang and Du, Bo},
  booktitle={Proceedings of the 32nd ACM International Conference on Multimedia},
  pages={4053--4062},
  year={2024}
}

@inproceedings{zhang2025viperson,
  title={VIPerson: Flexibly Generating Virtual Identity for Person Re-Identification},
  author={Zhang, Xiao-Wen and Zhang, Delong and Peng, Yi-Xing and Ouyang, Zhi and Meng, Jingke and Zheng, Wei-Shi},
  booktitle={Proceedings of the IEEE/CVF International Conference on Computer Vision},
  pages={23374--23384},
  year={2025}
}

@article{zheng2018cross,
  title={Cross-pose lfw: A database for studying cross-pose face recognition in unconstrained environments},
  author={Zheng, Tianyue and Deng, Weihong},
  journal={Beijing University of Posts and Telecommunications, Tech. Rep},
  volume={5},
  number={7},
  pages={5},
  year={2018}
}

@article{zheng2017cross,
  title={Cross-age lfw: A database for studying cross-age face recognition in unconstrained environments},
  author={Zheng, Tianyue and Deng, Weihong and Hu, Jiani},
  journal={arXiv preprint arXiv:1708.08197},
  year={2017}
}

@inproceedings{li2015generative,
  title={Generative moment matching networks},
  author={Li, Yujia and Swersky, Kevin and Zemel, Rich},
  booktitle={International conference on machine learning},
  pages={1718--1727},
  year={2015},
  organization={PMLR}
}

@article{robbins1951stochastic,
  title={A stochastic approximation method},
  author={Robbins, Herbert and Monro, Sutton},
  journal={The annals of mathematical statistics},
  pages={400--407},
  year={1951},
  publisher={JSTOR}
}

@article{wang2025scone,
  title={Scone: Bridging Composition and Distinction in Subject-Driven Image Generation via Unified Understanding-Generation Modeling},
  author={Wang, Yuran and Zeng, Bohan and Tong, Chengzhuo and Liu, Wenxuan and Shi, Yang and Ma, Xiaochen and Liang, Hao and Zhang, Yuanxing and Zhang, Wentao},
  journal={arXiv preprint arXiv:2512.12675},
  year={2025}
}

@article{lin2023privacy,
  title={Privacy-protected person re-identification via virtual samples},
  author={Lin, Yutian and Guo, Xiaoyang and Wang, Zheng and Du, Bo},
  journal={IEEE Transactions on Information Forensics and Security},
  volume={18},
  pages={5495--5505},
  year={2023},
  publisher={IEEE}
}

@inproceedings{jia2025balancing,
  title={Balancing Privacy and Performance: A Many-in-One Approach for Image Anonymization},
  author={Jia, Xuemei and Du, Jiawei and Wei, Hui and Xue, Ruinian and Wang, Zheng and Zhu, Hongyuan and Chen, Jun},
  booktitle={Proceedings of the AAAI Conference on Artificial Intelligence},
  volume={39},
  number={17},
  pages={17608--17616},
  year={2025}
}

@inproceedings{zhong2023refined,
  title={Refined semantic enhancement towards frequency diffusion for video captioning},
  author={Zhong, Xian and Li, Zipeng and Chen, Shuqin and Jiang, Kui and Chen, Chen and Ye, Mang},
  booktitle={Proceedings of the AAAI conference on artificial intelligence},
  volume={37},
  number={3},
  pages={3724--3732},
  year={2023}
}

@article{hu2025contrastive,
  title={Contrastive-Generative-Contrastive: Neutralize Subjectivity in Sketch Re-Identification},
  author={Hu, Zechao and Yang, Zhengwei and Li, Hao and Wang, Zheng},
  journal={IEEE Transactions on Information Forensics and Security},
  year={2025},
  publisher={IEEE}
}

@inproceedings{du2023minimizing,
  title={Minimizing the accumulated trajectory error to improve dataset distillation},
  author={Du, Jiawei and Jiang, Yidi and Tan, Vincent YF and Zhou, Joey Tianyi and Li, Haizhou},
  booktitle={Proceedings of the IEEE/CVF conference on computer vision and pattern recognition},
  pages={3749--3758},
  year={2023}
}

@article{du2024diversity,
  title={Diversity-driven synthesis: Enhancing dataset distillation through directed weight adjustment},
  author={Du, Jiawei and Zhang, Xin and Hu, Juncheng and Huang, Wenxing and Zhou, Joey T},
  journal={Advances in neural information processing systems},
  volume={37},
  pages={119443--119465},
  year={2024}
}
}


\end{document}